\documentclass[conference]{IEEEtran}
\IEEEoverridecommandlockouts
\usepackage{cite}
\usepackage{amsmath,amssymb,amsfonts}
\usepackage{algorithmic}
\usepackage{graphicx}
\usepackage{textcomp}
\usepackage{xcolor}
\usepackage{pifont}
\def\BibTeX{{\rm B\kern-.05em{\sc i\kern-.025em b}\kern-.08em
    T\kern-.1667em\lower.7ex\hbox{E}\kern-.125emX}}

\usepackage[absolute,showboxes]{textpos}
\TPMargin{5pt}
\newcommand{\copyrightstatement}{
    \begin{textblock}{0.84}(0.08,0.91)    % tweak here: {box width}(leftposition, rightposition)
         \noindent
         \footnotesize
Copyright 2019 IEEE. https://doi.org/10.1109/IJCNN.2019.8851907
Published in the 2019 International Joint Conference on Neural Networks (IJCNN), 14-19 July 2019 in Budapest, Hungary. Personal use of this material is permitted. However, permission to reprint/republish this material for advertising or promotional purposes or for creating new collective works for resale or redistribution to servers or lists, or to reuse any copyrighted component of this work in other works, must be obtained from the IEEE. Contact: Manager, Copyrights and Permissions / IEEE Service Center / 445 Hoes Lane / P.O. Box 1331 / Piscataway, NJ 08855-1331, USA. Telephone: + Intl. 908-562-3966.
    \end{textblock}
}
\begin{document}

\title{Underwater Fish Detection with Weak Multi-Domain 
 Supervision}

% 1\textsuperscript{st} 
\author{\IEEEauthorblockN{Dmitry A. Konovalov}
\IEEEauthorblockA{\textit{College of Science and Engineering} \\
\textit{James Cook University}\\
Townsville, Australia \\
dmitry.konovalov@jcu.edu.au}
\and
\IEEEauthorblockN{Alzayat Saleh}
\IEEEauthorblockA{\textit{College of Science and Engineering} \\
\textit{James Cook University}\\
Townsville, Australia \\
alzayat.saleh@my.jcu.edu.au}
\and
\IEEEauthorblockN{Michael Bradley}
\IEEEauthorblockA{\textit{College of Science and Engineering} \\
\textit{James Cook University}\\
Townsville, Australia \\
michael.bradley@my.jcu.edu.au}
\and
\IEEEauthorblockN{Mangalam Sankupellay}
\IEEEauthorblockA{\textit{College of Science and Engineering} \\
\textit{James Cook University}\\
Townsville, Australia \\
mangalam.sankupellay@jcu.edu.au}
\and
\IEEEauthorblockN{Simone Marini}
\IEEEauthorblockA{\textit{Institute of Marine Sciences} \\
\textit{National Research Council of Italy}\\
Forte Santa Teresa, 19032, La Spezia, Italy \\
simone.marini@sp.ismar.cnr.it}
\and
\IEEEauthorblockN{Marcus Sheaves}
\IEEEauthorblockA{\textit{College of Science and Engineering} \\
\textit{James Cook University}\\
Townsville, Australia \\
marcus.sheaves@jcu.edu.au}
}

% Simone MariniORCID: orcid.org/0000-0003-0665-7815
% National Research Council of Italy, Institute of Marine Sciences, Forte Santa Teresa, 19032, La Spezia, Italy
% Simone Marini, Emanuela Fanelli, Valerio Sbragaglia, Ernesto Azzurro, Joaquin Del Rio Fernandez and Jacopo Aguzzi contributed equally.

\maketitle
\copyrightstatement

\begin{abstract}
Given a sufficiently large training dataset, 
it is relatively easy to train a modern convolution 
neural network (CNN) as a required image classifier. 
However, for the task of fish classification and/or fish detection, 
if a CNN was trained to detect or classify particular fish species
in particular background habitats,
the same CNN exhibits much lower accuracy when applied to
new/unseen fish species and/or fish habitats. Therefore, in practice,
the CNN needs to be continuously fine-tuned to improve its classification accuracy to handle new 
project-specific fish species or habitats. In this work we present a labelling-efficient method
of training a CNN-based fish-detector 
(the Xception CNN was used as the base) 
on relatively small numbers (4,000) of project-domain 
underwater 
fish/no-fish images from 20 different habitats.
Additionally, 17,000 of known
negative (that is, missing fish) general-domain (VOC2012) 
above-water images were 
used. Two publicly available fish-domain datasets 
supplied additional 27,000 of 
above-water and underwater 
positive/fish images.
By using this multi-domain collection of images, the 
trained Xception-based binary (fish/not-fish) 
classifier achieved 
0.17\% false-positives and 0.61\% false-negatives on
the project's 20,000 negative 
and 16,000 positive holdout test images, respectively. 
The area under the ROC curve (AUC) was 99.94\%.

\end{abstract}


% Use "Eq" instead of "Equation" for equation citations.
\section{Introduction}

For the purpose of fish monitoring, 
remote underwater video (RUV) recording is a promising tool for fisheries, ecosystem management and conservation programs\cite{Shafait2016,SalmanSVM2017}. 
RUV applications are primarily divided into {\em baited}\cite{SalmanSVM2017} or {\em unbaited}. 
The focus of this study was the unbaited RUV processing because it uniquely 
offers the following  benefits: information about early life-history stages, and  
the spatial distribution and temporal dynamics of juveniles.
Such information is critical to fisheries and conservation management because it provides: 
(a) knowledge of juvenile habitats that need to be protected; (b) an understanding of the extent and direction of change of populations; (c) the ability to predict the size of future harvestable stocks; and (d) an understanding of the impact of habitat/environmental change on recruitment and survival through early life-history stages.

With the advent of consumer-grade action cameras, it is financially viable to deploy a large number of RUVs  especially within the recreational scuba diving 30-meter depth limit. 
However, the amount of data that need to be processed from the deployed RUVs can quickly overwhelm the resources of human video viewers, often rendering video analysis prohibitively costly.

\begin{figure}[htbp]
\centering
\includegraphics[width=0.48\textwidth]{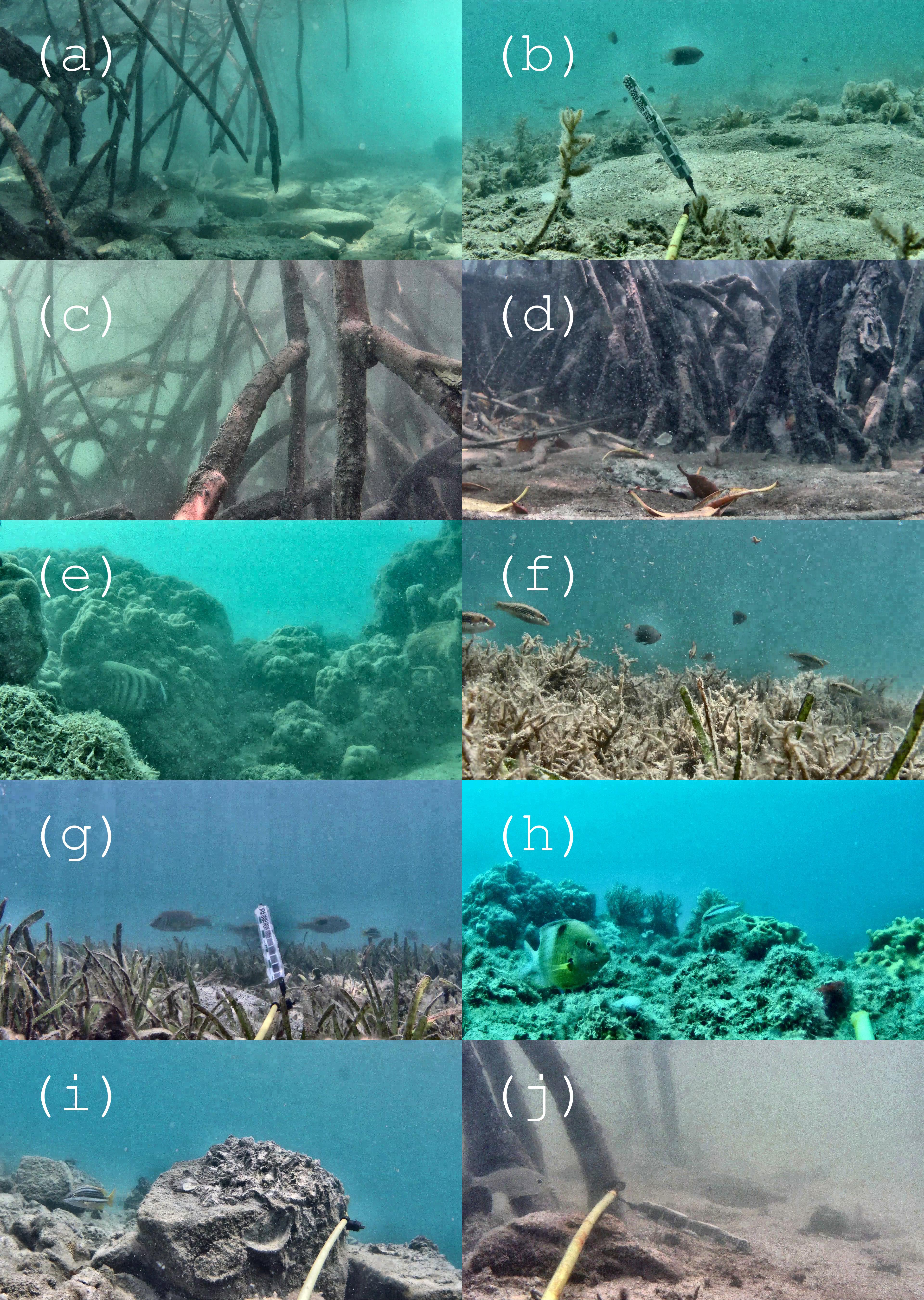}
\caption{Typical CLAHE processed fish-containing video frames from the first ten considered habitats.}
\label{fig1x10}
\end{figure}

\begin{figure}[htbp]
\centering
\includegraphics[width=0.48\textwidth]{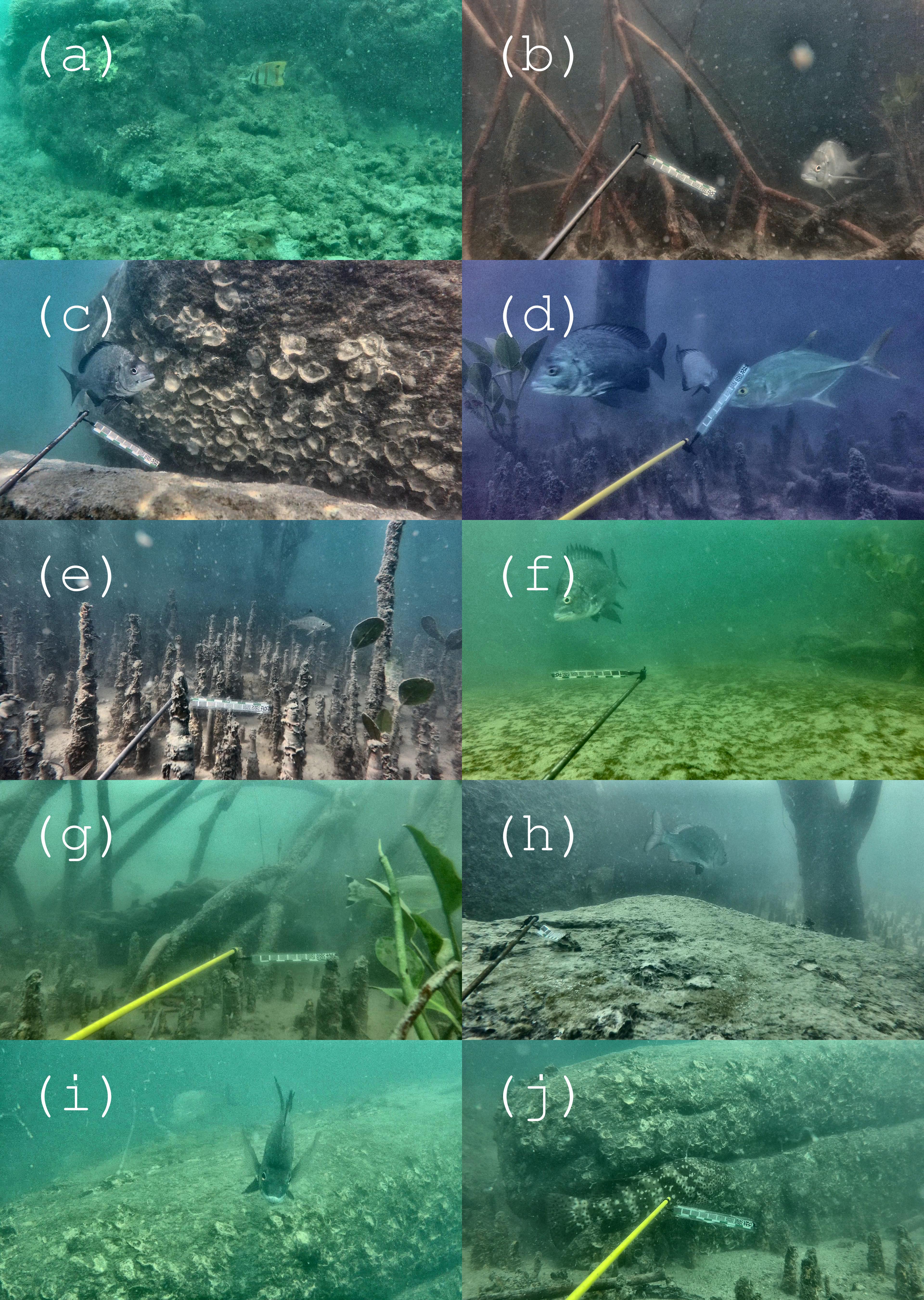}
\caption{Typical CLAHE processed fish-containing video frames from the 11th-20th habitats.}
\label{fig2x10}
\end{figure}

Conservation management requires unbaited RUVs to be placed in visually
complex underwater habitats (Figs.~\ref{fig1x10} and \ref{fig2x10}), 
where the traditional fish detection methods become 
unreliable (see Section~\ref{subsec:reatedWork}).
Modern Deep Learning \cite{LeCun15} convolutional neural networks (CNNs) are currently
achieving state of the art object-detection results 
in a wide variety of application domains; for example,  
automatic cattle detection from drones\cite{Rivas2018}.
Since the majority of unbaited videos do not contain any fish, 
the maximum positive impact could be achieved by using a CNN to  
automatically detect and discard the {\em empty} video clips/frames. 

In this study we developed a labelling-efficient procedure for
training a CNN-based \cite{Xception} binary image 
classifier (fish/non-fish) 
for fish-detection 
(see XFishHmMp in Section~\ref{sec:cnn})
and fish-localization (see XFishHm in Section~\ref{sec:location}).
The structure of this paper is as follows. 
Section~\ref{subsec:reatedWork} reviews recent development in 
underwater fish detection and classification. 
Section~\ref{subsec:dataset} describes the
labelling-efficient training and testing
data preparation protocol.
Section~\ref{sec:pipeline} presents the training pipeline.   
Section~\ref{sec:domain} introduces the main novel aspect of this work: weakly supervised 
training of the CNN fish-detector using external-to-project
image domains.
Section~\ref{sec:results} presents the results from 
the project test images, which were not used in the training of the fish-detector.

\subsection{Related Work}
\label{subsec:reatedWork}

The first large-scale automatic image and 
video-based fish detection and species classification study was the Fish4Knowledge (F4K) \cite{LifeCLEF2014,BOOM2014,f4k,f4kFinalReport} project, which was run over five years during 2010-2015.
F4K accumulated thousands of hours of underwater video clips of
coral reefs in Taiwan. 
Similar (to our work) studies since the F4K project are reviewed next.

The manually-annotated LCF-14 dataset of 30,000 fish images and 1,000 video clips 
containing ten fish species was reported by \cite{FishCLEF2014}. 
The images and videos were used as the challenge 
dataset for the fish task of the LifeCLEF2014 \cite{LifeCLEF2014} contest, and were derived from the F4K \cite{BOOM2014,f4k,f4kFinalReport} project. The VLfeat-BoW \cite{vedaldi08vlfeat,VLfeat2010} 
classification method was used as the baseline for the task of recognizing fish in still images achieving 97\% average precision ($AP$) and 91\% average recall (AR), defined as \cite{Salman2016} 
\begin{eqnarray}  
\label{eq:precision}
	AP = \frac{1}{c}\sum_{j=1}^c TP_j/(TP_j+FP_j),  
\end{eqnarray}
\begin{eqnarray}  
\label{eq:recall}
	AR = \frac{1}{c}\sum_{j=1}^c TP_j/(TP_j+FN_j), 
\end{eqnarray}
where $TP_j$, $FP_j$ and $FN_j$ are the numbers of true-positive, false-positive and 
false-negative classified results 
for the $j$th species, respectively, and where $c$ is the number of species.
For the videos, the ViBe \cite{ViBe} background subtraction algorithm was applied first, 
then followed by the VLfeat-BoW \cite{vedaldi08vlfeat,VLfeat2010} achieving only $AP=AR=54$\%. 
On the same LCF-14 test videos, very similar $AP \approx AR \approx 50$\% was also reported by 
\cite{Blanc2014} using a Support Vector Machine (SVM) classifier.
From (\ref{eq:precision}), the significantly lower recall value (91\%) compared to
the average precision (97\%) was due to the larger number of false-negatives $FN_j$ (compared to false-positives $FP_j$). 
Furthermore, the dramatically worse results ($AP \approx AR=50-54$\%) on videos
highlighted the need for more accurate fish {\em detection} methods 
\cite{FishCLEF2014,Blanc2014}, 
which is the main focus of this study.

The AlexNet CNN \cite{AlexNet} together with the 
Fast-R-CNN \cite{Girshick_2015_ICCV} method 
were applied in \cite{Li2015} to classify 12 different fish species in 24,272 images, which were a manually curated subset of the train and test images 
from the fish task competition of LifeCLEF2014 \cite{LifeCLEF2014,FishCLEF2014}. 
The LifeCLEF2014 fish task images were derived from the F4K \cite{BOOM2014} 
collection. 
Fast-R-CNN is a Fast Region-based Convolutional Neural Network
\cite{Girshick_2015_ICCV} method of object detection and classification in images.
A mean average precision (mAP) of 81.4\% was achieved \cite{Li2015} across the 12 considered species, where the mAP was defined as the total area under the precision-recall curve 
(see \cite{Li2015} for the exact definition). 
The most relevant aspect of \cite{Li2015} was that every train and test image 
was manually selected to contain one of the 12 species. Therefore, 
the method's ability to {\em detect} each species in the unconstrained underwater 
videos remained unknown.

The LCF-14 \cite{FishCLEF2014} 
dataset was used in \cite{Shafait2016} to create 
$32 \times 32$ gray training and test images.
Then, the face recognition algorithm of \cite{Hu2012} was applied 
to classify test images by finding the most 
similar species-specific images. 
Average classification accuracy of 94.6\% was reported, 
which was a significant improvement over the conceptually similar method of 
sparse image representation \cite{HSIAO201413}. 
However, the face-recognition \cite{Shafait2016} approach, and hence its accuracy, 
relied on an external method of \cite{Spampinato2008} 
for extracting and cropping fish sub-images from a given video.

An earlier version of the mixture of Gaussians (MoG) \cite{MOG2006}
algorithm was used in \cite{Spampinato2008} 
to segment moving fish from 
the stationary underwater background. 
The background-subtraction MoG 
\cite{MOG2006} algorithm works extremely well 
when the variations in the {\em background} 
pixel intensities are on average less than 
the variations due to the moving {\em foreground} object. 
For example, 
the clear and debris-free water at the top-right corner of
Fig~\ref{fig1x10}(a) or the top of Fig~\ref{fig1x10}(b).
The MoG is readily 
available in many common 
software packages such as Matlab and OpenCV \cite{itseez2017opencv}.
Unfortunately, 
the standard motion-based 
fish detection methods (for example, \cite{Spampinato2008}), by design, 
could not distinguish between floating debris and juvenile fish of comparable size, or
when the fish is stationary.
For example, the fish in the center of Fig~\ref{fig1x10}(j) is indistinguishable from 
the ground debris, 
and the fish in the left-middle (same sub-figure) 
remained stationary for many seconds.
Furthermore, the MoG-type \cite{MOG2006} methods fail when the background 
pixel variations are comparable with the slow-moving fish; 
for example, bottom-left {\em Lutjanus argentimaculatus} in Fig~\ref{fig1x10}(a)
or middle-left fish in Fig~\ref{fig1x10}(c).

The LifeCLEF2015 Fish classification challenge dataset 
LCF-15\cite{LifeCLEF2015,FishCLEF2015dataset} 
contained 20,000 labeled images, 
93 videos and 15 fish species. Both LCF-14 \cite{FishCLEF2014} 
and LCF-15 datasets were used in \cite{Salman2016},
where the videos were processed by extracting frames as separate images, 
then all available images were resized to 
$32 \times 32$ shape and converted to grayscale.
The use of only a three-layer CNN \cite{Salman2016} achieved $AP=97.18$\% when the CNN was trained on LCF-15 but tested on LCF-14. However, when trained on LCF-14 but tested on the noisy and poorer quality LCF-15, the CNN's performance degraded to $AP=65.36$\%. Furthermore, the same research group later reported \cite{SalmanSVM2017}
that the classification accuracy of the \cite{Salman2016}'s CNN degraded from $87.46$\% 
on the LCF-15 dataset to 
$53.5$\% on a completely different dataset\cite{SalmanSVM2017}.
This performance drop illustrates the technology challenge faced by any fish-monitoring fishery 
or ecology project: it is unknown how  
to pre-train a generic fish detection CNN and use it with confidence
in different environmental locations and/or to detect 
unknown (for the CNN) species.
Hence, the financial and human cost of setting up and training a project-specific 
CNN becomes a critical factor, which is a key issue our work is trying to address. 

Baited remote underwater cameras were used to collect videos 
from kelp, seagrass, sand and coral reef habitats in Western Australia\cite{HarveyPLOSone2013}. 
The videos were processed in \cite{SalmanSVM2017} to extract and label 2,209 images containing 16 fish species including {\em other} species as a separate 17th label. The images were re-sized and cropped to $224 \times 224 \times 3$ shape, where the three color channels were retained (hence the extra $\times 3$). 
Note the significant increase in image training complexity compared to the $32 \times 32 \times 1$ gray 
images of \cite{Salman2016,Shafait2016}. 
The following three CNN architectures were used in \cite{SalmanSVM2017}: AlexNet\cite{AlexNet}, VGGNet\cite{VGG16}, and ResNet\cite{ResNet}. 
The CNNs' original layers were initialized by loading the weights pre-trained on the ImageNet's \cite{imagenet_cvpr09} vast collection of images, which  
is commonly referred to as the {\em transfer learning} or {\em knowledge transfer} setup 
or technique\cite{Razavian2014,MidLevelTrans}.
An ImageNet-trained CNN often exhibits superior performance compared to the same but 
randomly initialized CNN, when the CNN is re-trained and/or re-purposed for different 
classes of images\cite{MidLevelTrans}; for example, the 16 fish species in \cite{SalmanSVM2017}.
The three considered ImageNet-trained CNNs were applied without further training to extract image features and then \cite{SalmanSVM2017} 
used the features as input into a standard SVM classifier.    
Out of the considered three CNNs, 
the ImageNet pre-trained ResNet\cite{ResNet} together with the SVM classifier achieved the best accuracy of $89$\% on the testing subset of 663 images. Furthermore, 
the ResNet+SVM combination achieved even better accuracy of $96.73$\% 
on the LCF-15 \cite{LifeCLEF2015,FishCLEF2015dataset} dataset. 
Note that similar to the preceding Fast-R-CNN  \cite{Li2015} and face-recognition-type \cite{Shafait2016} studies,
the ResNet+SVM \cite{SalmanSVM2017} focused on 
the classification of externally (and manually) detected
and appropriately cropped images. 
Therefore, the reported ResNet+SVM's 
high classification accuracy could only be achieved if 
it is accompanied by an automatic fish detection and bounding-box segmentation 
method of comparably high accuracy, 
which at present are estimated as only 
50\% accurate \cite{FishCLEF2014,Blanc2014}. 

Focusing only on the fish detection task, \cite{TrackingFishSciRep2018} reported 
a method of automatic fish counting in real-world videos, which is referred to as the 
OBSEA method hereafter. A binary (fish/no-fish) 
classifier was trained 
on 11,920 images collected at the OBSEA testing-site 
\cite{OBSEA} in 2012, and then tested on 10,961 imaged acquired at the same site in 2013. 
The OBSEA method consisted of two distinct steps. 
Within the first step, Regions of Interest (RoI) were automatically extracted from all training and test images. The RoI step used consecutive images sorted by the acquisition time and then essentially extracted the image differences as RoIs. Conceptually the RoI step is identical to the MoG-type  \cite{MOG2006} methods and therefore arguably 
would exhibit similar limitations: a large percentage of false-negatives when the fish 
is stationary, slow moving or below the adopted detection threshold. 
The figures and supplementary video of  \cite{TrackingFishSciRep2018}
clearly demonstrated this issue, 
where many fish instances were not segmented by the RoI step.
The second step of the OBSEA method applied a genetic programming method from \cite{Corgnati2016} to deliver a binary fish/not-fish classification for each of the 
segmented RoIs from the first step. Within a 10-fold cross-validation framework, the classifier achieved 92\% validation accuracy on manually labelled RoIs. 
Note that similar to the ResNet+SVM \cite{SalmanSVM2017} results, 
the reported accuracy can only be achieved on the per-fish/per-image basis if 
the preceding RoI or bounding-box segmentation step delivers appropriately low 
false-negative and false-positive rates, which was not reported in \cite{TrackingFishSciRep2018}. 

Based on this review of recent studies, 
the following working hypotheses were adopted for our study:
\begin{itemize}
	\item{Given a RoI or a bounding box in an image, there have been a number of 
methods achieving 85\%-95\%
accuracy of correctly classifying fish species or fish/not-fish detection.
}
	\item{All reviewed classifiers required 
	human-intensive annotation/labelling of 
RoI/bounding-boxes for each training image.
}
	\item{Trained classifiers are highly specialized to the training 
fish species and/or the training environmental habitat, and should not be 
assumed to work equally well on different species and/or different backgrounds.
}
\item{The accuracy of automatic fish-related RoI/segmentation methods is highly dependent
on the image/video background/habitat.
}
\item{In complex reef-type habitats, the RoI extraction methods are only 50\%-80\%
accurate, which is significantly less accurate than the classifiers
from the 
corresponding studies.
}
\end{itemize}

\section{Materials and methods}

\begin{figure}[htbp]
\centering
\includegraphics[width=0.45\textwidth]{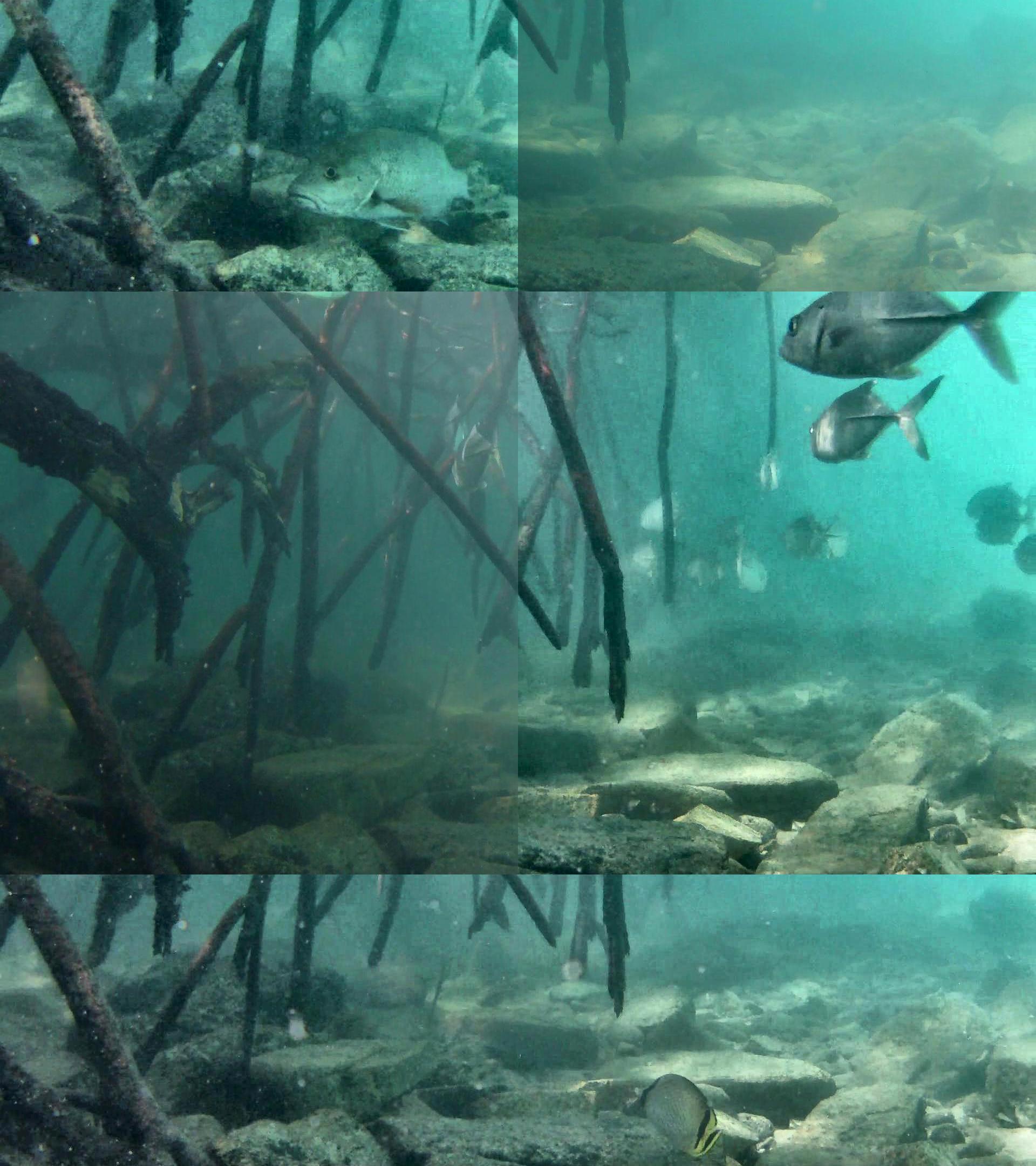}
\caption{Examples of frames from mangroves habitat with:
one {\em Lutjanus argentimaculatus} adult (top row); one {\em Chaetodon vagabundus} (bottom row); and
multiple {\em Caranx sexfasciatus} juveniles (middle row). 
Top-left, middle-right and bottom-row sub-images were histogram equalized via CLAHE\cite{CLAHE}.}
\label{fig:CLAHE}
\end{figure}

\subsection{Labelling-Efficient 
Dataset Preparation Protocol}
\label{subsec:dataset}

The essential goal of this study was to design 
and test a practical and labelling-efficient 
data preparation protocol, which could be used
in future fish-survey studies. The following protocol was utilized 
with realistic estimations of the required human labor for 
project planning and costing.

Video clips from 20 diverse habitats were selected 
(see typical examples in Figs.~\ref{fig1x10} and \ref{fig2x10}).
Video clips were recorded in a range of different environmental conditions present across a near-shore island chain of the Great Barrier Reef (Palm Islands, Queensland, Australia). These clips represent the range of different conditions encountered during a tropical marine fish survey, and form part of the field data presented in an assessment of juvenile fish habitat \cite{Bradley2019}. 
Recording sites varied in three-dimensional habitat architecture, levels of natural shading, current and wave energy levels, levels of suspended sediments, organic flocculation (marine snow), turbidity and salinity.

The video clips were visually examined to determine if they contained at least one fish. 
Then all clips containing fish were placed into 
sub-folder named {\em valid}, while all the clips without any fish species
were placed into the sub-folder named {\em empty}.
All but one habitat (Fig.~\ref{fig1x10}b) 
had at least one valid and one empty clip, 
where    
the collected valid and empty clips could be reused in
the future projects to gradually build a more comprehensive
fish-detection training dataset. 
This sorting took approximately two days (10 hours) for an experienced 
marine biologist already familiar with the content of the videos. 
All clips were then converted to individual frame images where the first, 11th, 21st, etc. 
frames (intervals of 10) were saved for
training (denoted as FD10) 
and the remaining frames were saved separately for testing 
(denoted as FD10-Test).
In total the clips yielded 40,000 frames, where 
the FD10 dataset  
contained 1764 positive (fish) 
and 2253 negative (no-fish) images. 
The FD10-Test dataset  contained 16,000 positive and 
20,000 negative images. This clip-level labelling 
is highly human-labor efficient in generating thousands of 
project-specific image-level annotations. 
However, the proposed labelling procedure
is valid only if both fish and no-fish 
clips are available from the same location. A CNN model 
needs to learn the fish features 
(only present in positive clips) 
and learn to ignore the underwater habitat features 
(available in both negative and positive clips).

The FD10 and FD10-Test 
collections of the unprocessed original frames
were further processed by the CLAHE \cite{CLAHE}
algorithm from the OpenCV \cite{itseez2017opencv} library
to create the corresponding FD10c and FD10c-Test 
datasets.
The CLAHE default clip limit value was retained at 2, and 
the CLAHE tile grid sizes were set at 
16 column tiles and 8 row tiles. 
Each image was first converted from the 
RGB color space to the CIELAB
\cite{CIELAB}
space via the relevant OpenCV function, 
where the luminosity ($L$) channel was then processed by 
the CLAHE algorithm. Visually, the CLAHE processed frames 
were significantly better than the unprocessed raw video frames 
(see the effect of CLAHE in Fig.~\ref{fig:CLAHE}).

The original $1080 \times 1920$ ($rows \times columns$) resolution
project videos (not the training clips) 
were approximately 1TB
in combined disk storage size 
and 200 hours in total duration time (600 videos, each 20 min long), 
which required at least 200
hours of paid (or effectively paid via the lost opportunity-cost) processing by marine biologists. 
More than half of the video frames did not contain any fish species, 
which meant at least 100 human hours were wasted analyzing the original videos. 
The unproductive human effort was further compounded by searching within the videos for the relevant fish-containing sections of interest. Since such fish surveys are repeated regularly, the presented fish-detection method and procedures 
could be further refined in future studies.

\subsection{Project-Domain}
\label{sec:cnn}
For the supervised training, the most 
human-efficient labelling is achieved by the image-level
class labels \cite{WhatIsThePoint2014}.
If annotated by a class label, 
one or more instances of the class are curated to be 
present somewhere in the image scene, directly visible or directly 
inferrable. 

Out of the common ImageNet-trained and 
available in Keras \cite{keras} CNNs,
Xception \cite{Xception} was selected as the base CNN.
The required binary fish/no-fish classifier (denoted as XFishMp)
was constructed
by replacing the Xception's 1,000-class top with one spatial/global
maximum pooling layer (hence the ``Mp'' abbreviation in XFishMp) 
followed by a 0.5-probability dropout layer, and then 
by a one-class dense layer with the {\em sigmoid} activation function.
The Xception-based XFishMp contained the smallest number of
trainable 
parameters (20.8 million ) 
compared to 23.5 million in the ResNet50-based and 21.7 million in the InceptionV3-based XFish's equivalents. 
Another CNN configuration was also considered and denoted XFishHmMp, 
where the XFishMp's global max-pool was moved to be the last layer 
and the one-class dense layer was converted to 
a convolution layer. The ``Hm'' naming mnemonic was due to 
the one-class convolution layer yielding a two-dimensional
{\em heatmap} of $[0,1]$-ranged values. 

In this study, the FD10 dataset consisted of 4017 color images 
(1764 with fish and 2253 without fish), where 
each image had 1080 pixel rows and 1920 columns
($1080 \times 1920$ shape).
The fish sizes were mostly within the $[30,300]$-pixel range. 

In comparison with 
the ImageNet's more than one million images (used to train 
Xception), the FD10 dataset is very small (4017 images). 
Hence, additional measures were required to prevent over-fitting
of the XFish CNNs. 
The first such measure was to use the training color images
only via their grayscale versions.  
As the color variation in reef fish species is generally greater 
than the variation in fish shapes, by removing the color features,
the XFish CNNs were 
potentially more likely to learn ({\em generalize})
the fish shapes, rather than to memorize ({\em fit}) pixel colors.
Furthermore, the underwater background colors vary dramatically 
(Figs.~\ref{fig1x10} and \ref{fig2x10}) and therefore 
the considered FD10's 4017 images could be classified
more easily if the color channels were included. However, 
such fitting of the colors would have little or no generalization
value beyond the considered training dataset.

Since the ImageNet-trained Xception required the three color
channels in its input, a trainable gray-to-RGB 
convolution conversion layer was added to the front of the 
XFish CNNs
to accept the one-channel grayscale training images. 

In order to achieve practical training times and to fit the training
onto common GPUs (Nvidia GTX 1070 and GTX 1080 Ti were used), the 
training image dimensions were limited 
to the $512 \times 512$ shape. Then, in addition to
the grayscale input images, the following augmentations were
performed to reduce over-fitting (that is, additional regularization).
Each original $1080 \times 1920$
image was converted to the grayscale and then zero padded 
by a 5\% border yielding $1188 \times 2112$ shaped images.
The padded images were downsized 
(or zero-padded if required in Section~\ref{sec:domain}) 
to the $512 \times 512$ shape 
and then:
\begin{itemize}
    \item randomly rotated within  $[-20, 20]$ degree range;
    \item randomly flipped horizontally with the 0.5 probability;
    \item their rows and columns were resized
 independently by random scales from $[0.9, 1]$ range;
    \item after zero-padding to the $512 \times 512$-shape,
random perspective transformation was applied;
    \item normal Gaussian noise was added and the final image values
were clipped to $[0,255]$ range, 
where the noise mean was zero, and the noise standard deviation
was randomly selected from $[0, 8]$ range;
\item the grayscale $[0,255]$-range pixel values were 
normalized to zero minimum and maximum of one with each image.
\end{itemize}

\subsection{Training Pipeline}
\label{sec:pipeline}

All considered models were trained in Keras \cite{keras} 
with the Tensorflow \cite{tensorflow2015-whitepaper} back-end,
where the Adam \cite{Adamax14} algorithm was used as the training optimizer. 
The Adam's initial learning-rate ($lr$) was set to $lr=1\times 10^{-5}$ for training XFishMp and to $1\times 10^{-4}$ for XFishHmMp, 
where the rate was halved every time the epoch {\em validation}
accuracy did not increase after 10 epochs.
The training was done in batches of four images 
and was restarted twice from the highest-accuracy model 
if the validation accuracy did not increase after 32 epochs, where
at each restart the initial $lr$ was multiplied by $0.9$. 
The validation subset of images was not augmented but only
pre-processed: 5\% zero-padded, resized to the $512 \times 512$ input shape, and normalized to the $[0,1]$ range. 

\subsection{Weak Supervision by External Domains}
\label{sec:domain}
Arguably, the only reason to use the project-domain
datasets is the absence of public fish-domain
image/video datasets of the required 
fish-species and of required image quality and quantity.
However, there are many general-domain image datasets where fish instances are labelled 
(for example, ImageNet \cite{imagenet_cvpr09}) 
or known to be missing (for example, VOC2012 \cite{VOC2012}).

The Xception CNN utilized here was trained on more 
than one million ImageNet images (including some fish images).
The project's FD10 collection of 4,000 training images was still very small 
for the modern high capacity CNNs, such as Xception.
Therefore, in this study, we regularized
XFish CNNs by 
using negative (not-fish) general-domain images, where
the 17,000 VOC2012 \cite{VOC2012} images were
used in this study to achieve weak negative supervision.
All of the original videos (used as the base
for this project's training clips) contained the above-water sections
at the beginning of each video, when the camera was manually 
turned on before being lowered to its underwater destination. 
The negative every-day type
VOC2012 images
assisted in more robust rejection of the
above-water false-positives.

For the weak positive supervision, two 
specialized fish-domain datasets were utilized:
the LCF-15 classification challenge dataset 
\cite{LifeCLEF2015,FishCLEF2015dataset} 
with 22.4 thousand fish images, and
the QUT2014 dataset 
\cite{QUT_fish_data,QUT2014}
with 4.4 thousand fish images.

In order to retain the weak nature of the
external mutli-domain datasets, we proposed the following
training pipeline. 
The FD10 was enlarged by total of 4,000 images
(denoted as FD10-VLQ), where
2,000 images were randomly selected from VOC2012 (automatically labeled as negative/no-fish) and
1,000 images from each LCF-15 and QUT2014 
(automatically labeled as positive/fish).
Then, the new 8,000 large FD10-VLQ dataset was
split 80/20\% into the training/validation subsets. 
Since many more images remained available in all three
considered domain-level datasets, at each training epoch, all 
4,000 additional external-domain images were
randomly re-drawn from their corresponding datasets.

\subsection{Fish Localization}
\label{sec:location}
Fish detection normally implies {\em localization} of the detected
fish within an image. XFishHmMp could be easily
converted for the localization task,
by removing its last max-pooling layer arriving at the XFishHm CNN.
XFishHm outputs a grayscale heatmap of the input image spatially downsized by 32, 
that is, the $512 \times 512$ grayscale image is converted into $16 \times 16$ heatmap of $[0,1]$-ranged values.
Weak localization supervision was achieved by 
deliberately (and human-time efficiently) selecting the 
fish-containing and missing-fish FD10 
video clips from the same underwater locations.
Note that due to their higher labelling costs, 
the direct fish-level supervision 
via, for example, bounding-box\cite{WhereAreTheBlobs2018}, 
pixel-level {\em semantic segmentation} 
or point-level \cite{WhatIsThePoint2014} annotations 
were considered outside the scope of this study.

\section{Results and Discussion}
\label{sec:results}

\subsection{Baseline}
To establish the baseline, XFishMp and XFishHmMp were
trained in Keras \cite{keras} 
with the Tensorflow \cite{tensorflow2015-whitepaper}
back-end on the identical random (controlled by a fixed seed value)
train/validation split of the FD10 and FD10c 
datasets, where the 
label-stratified split was 80\% for training and 20\% for validation.
The binary cross-entropy was used as the training loss. 

All FD10 and FD10c 
trained models (Table~\ref{tab1}) were applied to the 
FD10-Test dataset, which was not processed by CLAHE \cite{CLAHE}. 
On NVIDIA GTX 1080 Ti, the networks processed the test images
at 7-8 images per second (one image per batch), 
which was borderline acceptable 
for processing the large volume of underwater videos
in deployment,
for example, by further optimization of running in larger batches and/or 
only loading every second or third frames. 
However, additional CLAHE pre-processing reduced the testing 
rate to 
0.5-1 images per second and therefore was not considered 
as a currently-viable deployment option.

Since every 10th frame was used for training (or validation), it was
reasonable to expect that the remaining test frames (from the holdout FD10-Test dataset) would be classified exactly (zero false-negatives and zero false-positives). 
The default 0.5 threshold was used to accept the CNN activation
output as positive/fish, and classify as negative/no-fish
if the output value was less than the threshold. The lowest 
baseline false-positive rate ($FP/N=0.25$\%) 
was achieved by XFishMp (trained on FD10), see Table~\ref{tab1},
while lowest baseline false-negative rate ($FN/P=0.84$\%) 
was by the heatmap-based XFishHmMp
(trained on FD10). In Table~\ref{tab1}, $N$ and $P$ denoted 
the total number of negative and positive test images, respectively.

The training on the {\em cleaner} CLAHE-processed
FD10c images, reduced the CNN's generalization ability,
where the best baseline 
false-positive rate deteriorated from $FP/N=0.25$\% 
to 2.39\% for the XFishMp+FD10c CNN. 
A conceptually similar result was reported
by \cite{Salman2016}, 
where training on the noisy LCF-15 dataset achieved
higher accuracy (tested on cleaner LCF-14) 
than training on clean LCF-14 and testing on noisy LCF-15. 
Therefore, 
while visually appealing, 
the image-cleaning pre-processing is not necessary and could
even be detrimental to the CNN performance.

\begin{table}[htbp]
\caption{Confusion Matrix for the FD10-Test Dataset}
\begin{center}
\begin{tabular}{|c|l|l|l|}
\hline
\textbf{ }&
\multicolumn{1}{|c|}{\textbf{Model}} &
\multicolumn{2}{|c|}{\textbf{Predicted}} 
\\
\cline{2-4} 
\textbf{Actual} & \textbf{\textit{Train dataset}}
& \textbf{\textit{Negative}}&
\textbf{\textit{Positive}}
\\
\hline
\textbf{\textit{Negative}}  &  \textbf{XFishMp}
& \textbf{\textit{TN}} 
& \textbf{\textit{FP (FP/N\%)}}  \\
\textbf{\textit{(no-fish)}} 
 & FD10c & 19,623 & 481 (2.39\%) \\
 & FD10c-VLQ   
 & 19,849   & 255 (1.27\%)   \\
\textbf{\textit{N=20,104}}  
 & FD10 & 20,053 & 51 (0.25\%)  \\
 & FD10-VLQ & 20,005 & 99 (0.49\%) \\
\cline{2-4} 
 & \textbf{XFishHmMp} &  &  \\
 & FD10c & 18,378 & 1,726 (8.58\%) \\
 & FD10c-VLQ   & 18,638   
 & 1,466 (7.29\%)   \\
 & FD10 & 19,998 & 106 (0.53\%)  \\
 & FD10-VLQ   & 20,070   
 & 34 ({\bf 0.17\%})   \\
\hline
\textbf{\textit{Positive}}  & \textbf{XFishMp} & 
\textbf{\textit{FN (FN/P\%)}} 
& \textbf{\textit{TP (AUC\%)}}  \\
\textbf{\textit{(fish)}} & FD10c & 876 (5.28\%) 
& 14,725 (99.24\%) \\
 & FD10c-VLQ   & 884 (5.32\%) 
 & 14,717 (99.31\%) \\
\textbf{\textit{P=16,601}} 
 & FD10 & 162 (0.98\%) & 15,439 (99.92\%) \\
 & FD10-VLQ    
 & 117 (0.71\%)    & 15,484  (99.96\%)     \\
\cline{2-4} 
 & \textbf{XFishHmMp} &  &   \\
 & FD10c & 1,713 (10.32\%) & 13,888 (96.48\%) \\
 & FD10c-VLQ   & 1,564 (9.42\%)   
 & 14,037 (96.90\%)   \\
 & FD10 & 139 (0.84\%) & 15,462 (99.92\%) \\
 & FD10-VLQ    & 101 ({\bf 0.61\%})    
 & 15,500 (99.94\%)   \\
\hline
\end{tabular}
\label{tab1}
\end{center}
\end{table}

% \ding{55}  <- x

\subsection{Multi-Domain Image-Level Supervision}

To observe the effect of the additional domain-level weak
supervision, the baseline-trained XFishMp and XFishHmMp CNNs 
were fine-tuned 
on the FD10-VLQ and FD10c-VLQ dataset (see Section~\ref{sec:domain}).
Note that the CLAHE pre-processing was not applied to any of the 
external images.
The training pipeline (Section~\ref{sec:pipeline}) remained nearly identical, where 
the corresponding starting 
learning rates were reduced by the factor of 10, 
and only one training cycle was used.
This means the training was not restarted once aborted.

% see ticks \ding{51} in 
The weak supervision by external images  
improved all of the baseline cases (Table~\ref{tab1}) to some degree. 
The heatmap-based XFishHmMp CNN (trained on raw FD10-VLQ)
achieved
the lowest possible false-positive ($FP/N=0.17$\%) and 
false-negative ($FN/P=0.61$\%) rates.
Only two cases did not improve with the selected {\em default}
detection threshold of $0.5$: 
false-positives ($FP$) of XFishMp(FD10-VLQ)
and false-negatives ($FN$) of XFishMp(FD10c-VLQ).
However, the receiver operating characteristics' (ROC) area under the curve (AUC)
\cite{ROC06} revealed 
that even in the two cases, a better separation of positive and negative activation values was achieved
(see the bottom-right sub-column of Table~\ref{tab1}).

% were very minor: $FP$ increased to 99 (from 51) 
% out of 20,104 negatives,
% and $FN$ increased to 884 (from 876) out of 16,601 positives.

Clearly, the additional positive weak supervision could not 
improve the false-negative rate ($FN$) significantly, where the external 
fish images (in LCF-15  \cite{LifeCLEF2015,FishCLEF2015dataset} 
and QUT2014 \cite{QUT_fish_data,QUT2014}) 
were very different from the project-domain fish images.
The external negative weak supervision was more likely to improve
the false-positive rate ($FP$), which indeed decreased   
in XFishMp(FD10c) from 481 to 255 (Table~\ref{tab1}).

\begin{figure}[htbp]
\centering
\includegraphics[width=0.48\textwidth]{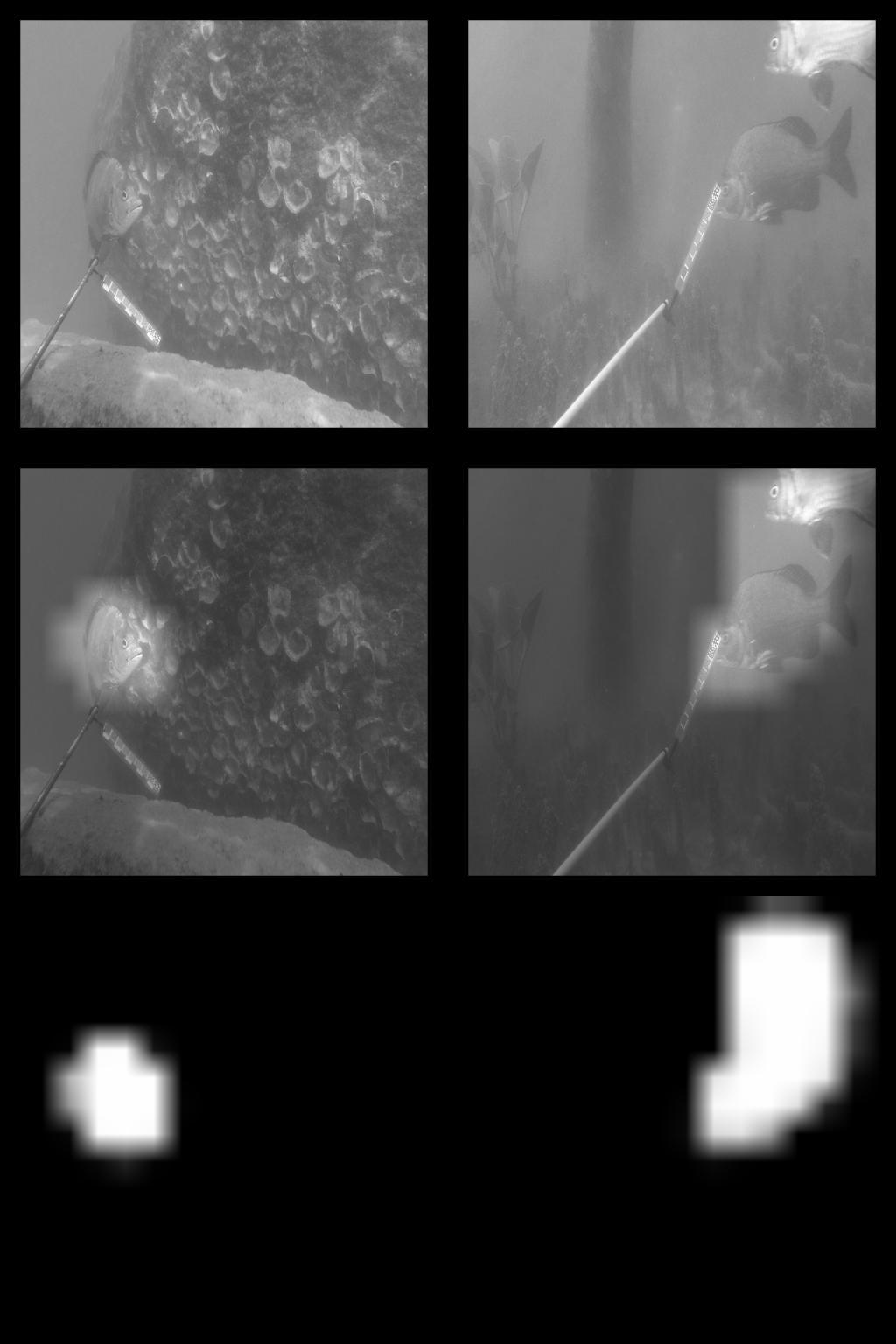}
\caption{Typical examples of fish 
correctly detected and localized by XFishHm (trained as XFishHmMp), 
where the top subfigures are the padded and re-scaled original grayscale images, 
middle are the images overlapped with the prediction heatmaps, 
and the bottom row are the prediction heatmaps.}
\label{fig:fish_seg}
\end{figure}

\subsection{Weakly Supervised Localization}
The heatmap-based XFishHmMp CNN 
achieved the lowest final $FP$ and $FN$ errors (Table~\ref{tab1}).
After removing the last max-pooling later, XFishHmMp 
was converted to the localization XFishHm CNN (see 
Section~\ref{sec:location}). Detailed analysis of the 
localization accuracy was left to future work
as it required the ground-truth bounding-boxes or segmentation 
masks for the FD10-Test images. Nevertheless, 
XFishHm was applied to all FD10 images
and results were visually inspected to verify
good consistency of the heatmap fish localization. Typical 
heatmap segmented examples are presented in Fig.~\ref{fig:fish_seg}, 
where the heatmaps were re-scaled 
to the original training size of $512 \times 512$ 
(from the XFishHm $16 \times 16$ output).

Since the XFishMp architecture did not exhibit consistently superior 
accuracy compared to XFishHmMp (Table~\ref{tab1}), 
and XFishMp could not be instantly converted to output the heatmaps,
we accepted XFishHmMp as the starting architecture for future work. 
Note, that 
the Xception CNN base in 
XFishHmMp could be trivially replaced by any other modern CNN, 
where any required 
input image normalization is automatically taken care by 
the trainable gray-to-RGB conversion layer.

\section{Conclusion}
In conclusion, 
we developed a novel training procedure 
for a relatively small number of 
project-domain images
to be utilized more effectively
when training a project-specific CNN fish-detector
together with a much larger pool of multi-domain images.
A human-time efficient labelling procedure was 
successfully tested. 
The regularizing effect of the weak supervision by external large
multi-domain image collections was verified. 
Pre-processing image cleaning
could reduce the model generalization performance.

% \section*{Acknowledgments}
% TODO. Any suggestions???

% \section*{References}
% \IEEEtriggeratref{10}
\bibliographystyle{IEEEtran}
\bibliography{IEEEabrv,fish2018}

\begin{thebibliography}{10}
\providecommand{\url}[1]{#1}
\csname url@samestyle\endcsname
\providecommand{\newblock}{\relax}
\providecommand{\bibinfo}[2]{#2}
\providecommand{\BIBentrySTDinterwordspacing}{\spaceskip=0pt\relax}
\providecommand{\BIBentryALTinterwordstretchfactor}{4}
\providecommand{\BIBentryALTinterwordspacing}{\spaceskip=\fontdimen2\font plus
\BIBentryALTinterwordstretchfactor\fontdimen3\font minus
  \fontdimen4\font\relax}
\providecommand{\BIBforeignlanguage}[2]{{%
\expandafter\ifx\csname l@#1\endcsname\relax
\typeout{** WARNING: IEEEtran.bst: No hyphenation pattern has been}%
\typeout{** loaded for the language `#1'. Using the pattern for}%
\typeout{** the default language instead.}%
\else
\language=\csname l@#1\endcsname
\fi
#2}}
\providecommand{\BIBdecl}{\relax}
\BIBdecl

\bibitem{Shafait2016}
F.~Shafait, A.~Mian, M.~Shortis, B.~Ghanem, P.~F. Culverhouse, D.~Edgington,
  D.~Cline, M.~Ravanbakhsh, J.~Seager, and E.~S. Harvey, ``Fish identification
  from videos captured in uncontrolled underwater environments,'' \emph{ICES
  Journal of Marine Science}, vol.~73, pp. 2737--2746, 2016.

\bibitem{SalmanSVM2017}
S.~A. Siddiqui, A.~Salman, M.~I. Malik, F.~Shafait, A.~Mian, M.~R. Shortis, and
  E.~S. Harvey, ``Automatic fish species classification in underwater videos:
  exploiting pre-trained deep neural network models to compensate for limited
  labelled data,'' \emph{ICES Journal of Marine Science}, vol.~75, pp.
  374--389, 2018.

\bibitem{LeCun15}
Y.~LeCun, Y.~Bengio, and G.~Hinton, ``Deep learning,'' \emph{Nature}, vol. 521,
  pp. 436--444, 2015.

\bibitem{Rivas2018}
A.~Rivas, P.~Chamoso, A.~González-Briones, and J.~M. Corchado, ``Detection of
  cattle using drones and convolutional neural networks,'' \emph{Sensors},
  vol.~18, no.~7, 2018.

\bibitem{Xception}
F.~Chollet, ``Xception: Deep learning with depthwise separable convolutions,''
  in \emph{2017 IEEE Conference on Computer Vision and Pattern Recognition
  (CVPR2017)}, 2017.

\bibitem{LifeCLEF2014}
A.~Joly, H.~Go{\"e}au, H.~Glotin, C.~Spampinato, P.~Bonnet, W.-P. Vellinga,
  R.~Planque, A.~Rauber, R.~Fisher, and H.~M{\"u}ller, ``Lifeclef 2014:
  Multimedia life species identification challenges,'' in \emph{Information
  Access Evaluation. Multilinguality, Multimodality, and Interaction}, ser.
  Lecture Notes in Computer Science, E.~Kanoulas, M.~Lupu, P.~Clough,
  M.~Sanderson, M.~Hall, A.~Hanbury, and E.~Toms, Eds., vol. 8685.\hskip 1em
  plus 0.5em minus 0.4em\relax Cham: Springer International Publishing, 2014,
  pp. 229--249.

\bibitem{BOOM2014}
B.~J. Boom, J.~He, S.~Palazzo, P.~X. Huang, C.~Beyan, H.-M. Chou, F.-P. Lin,
  C.~Spampinato, and R.~B. Fisher, ``A research tool for long-term and
  continuous analysis of fish assemblage in coral-reefs using underwater camera
  footage,'' \emph{Ecological Informatics}, vol.~23, pp. 83 -- 97, 2014,
  special Issue on Multimedia in Ecology and Environment.

\bibitem{f4k}
\BIBentryALTinterwordspacing
R.~B. Fisher \emph{et~al.} [Online]. Available: \url{www.fish4knowledge.eu}
\BIBentrySTDinterwordspacing

\bibitem{f4kFinalReport}
\BIBentryALTinterwordspacing
------. [Online]. Available: \url{https://bit.ly/2Ex7dnZ}
\BIBentrySTDinterwordspacing

\bibitem{FishCLEF2014}
C.~Spampinato, S.~Palazzo, P.~H. Joalland, S.~Paris, H.~Glotin, K.~Blanc,
  D.~Lingrand, and F.~Precioso, ``Fine-grained object recognition in underwater
  visual data,'' \emph{Multimedia Tools and Applications}, vol.~75, pp.
  1701--1720, 2016.

\bibitem{vedaldi08vlfeat}
\BIBentryALTinterwordspacing
A.~Vedaldi and B.~Fulkerson, ``{VLFeat}: An open and portable library of
  computer vision algorithms,'' 2008. [Online]. Available:
  \url{http://www.vlfeat.org}
\BIBentrySTDinterwordspacing

\bibitem{VLfeat2010}
------, ``{VLfeat}: An open and portable library of computer vision
  algorithms,'' in \emph{Proceedings of the 18th ACM International Conference
  on Multimedia}, ser. MM '10.\hskip 1em plus 0.5em minus 0.4em\relax New York,
  NY, USA: ACM, 2010, pp. 1469--1472.

\bibitem{Salman2016}
A.~Salman, A.~Jalal, F.~Shafait, A.~Mian, M.~Shortis, J.~Seager, and E.~Harvey,
  ``Fish species classification in unconstrained underwater environments based
  on deep learning,'' \emph{Limnology and Oceanography: Methods}, vol.~14, pp.
  570--585, 2016.

\bibitem{ViBe}
O.~Barnich and M.~V. Droogenbroeck, ``{ViBe}: A universal background
  subtraction algorithm for video sequences,'' \emph{IEEE Transactions on Image
  Processing}, vol.~20, no.~6, pp. 1709--1724, 2011.

\bibitem{Blanc2014}
K.~Blanc, D.~Lingrand, and F.~Precioso, ``Fish species recognition from video
  using svm classifier,'' in \emph{Proceedings of the 3rd ACM International
  Workshop on Multimedia Analysis for Ecological Data}, ser. MAED '14.\hskip
  1em plus 0.5em minus 0.4em\relax New York, NY, USA: ACM, 2014, pp. 1--6.

\bibitem{AlexNet}
A.~Krizhevsky, I.~Sutskever, and G.~E. Hinton, ``Imagenet classification with
  deep convolutional neural networks,'' in \emph{Proceedings of the 25th
  International Conference on Neural Information Processing Systems - Volume
  1}, ser. NIPS'12.\hskip 1em plus 0.5em minus 0.4em\relax USA: Curran
  Associates Inc., 2012, pp. 1097--1105.

\bibitem{Girshick_2015_ICCV}
R.~Girshick, ``Fast r-cnn,'' in \emph{The IEEE International Conference on
  Computer Vision (ICCV)}, December 2015.

\bibitem{Li2015}
X.~Li, M.~Shang, H.~Qin, and L.~Chen, ``Fast accurate fish detection and
  recognition of underwater images with fast r-cnn,'' in \emph{OCEANS 2015 -
  MTS/IEEE Washington}, 2015, pp. 1--5.

\bibitem{Hu2012}
Y.~Hu, A.~S. Mian, and R.~Owens, ``Face recognition using sparse approximated
  nearest points between image sets,'' \emph{IEEE Transactions on Pattern
  Analysis and Machine Intelligence}, vol.~34, pp. 1992--2004, 2012.

\bibitem{HSIAO201413}
Y.-H. Hsiao, C.-C. Chen, S.-I. Lin, and F.-P. Lin, ``Real-world underwater fish
  recognition and identification, using sparse representation,''
  \emph{Ecological Informatics}, vol.~23, pp. 13 -- 21, 2014, special Issue on
  Multimedia in Ecology and Environment.

\bibitem{Spampinato2008}
C.~Spampinato, Y.-H. Chen-Burger, G.~Nadarajan, and R.~B. Fisher, ``Detecting,
  tracking and counting fish in low quality unconstrained underwater videos,''
  in \emph{Proceedings of 3rd International Conference on Computer Vision
  Theory and Applications (VISAPP)}, vol.~2, 2008, pp. 514--519.

\bibitem{MOG2006}
Z.~Zivkovic and F.~van~der Heijden, ``Efficient adaptive density estimation per
  image pixel for the task of background subtraction,'' \emph{Pattern
  Recognition Letters}, vol.~27, pp. 773 -- 780, 2006.

\bibitem{itseez2017opencv}
Itseez, ``Open source computer vision library,''
  \url{https://github.com/itseez/opencv}, 2017.

\bibitem{LifeCLEF2015}
A.~Joly, H.~Go{\"e}au, H.~Glotin, C.~Spampinato, P.~Bonnet, W.-P. Vellinga,
  R.~Planqu{\'e}, A.~Rauber, S.~Palazzo, B.~Fisher, and H.~M{\"u}ller,
  ``Lifeclef 2015: Multimedia life species identification challenges,'' in
  \emph{Experimental IR Meets Multilinguality, Multimodality, and Interaction},
  ser. Lecture Notes in Computer Science, J.~Mothe, J.~Savoy, J.~Kamps,
  K.~Pinel-Sauvagnat, G.~Jones, E.~San~Juan, L.~Capellato, and N.~Ferro, Eds.,
  vol. 9283.\hskip 1em plus 0.5em minus 0.4em\relax Cham: Springer
  International Publishing, 2015, pp. 462--483.

\bibitem{FishCLEF2015dataset}
\BIBentryALTinterwordspacing
``Fish species recognition.'' [Online]. Available: \url{https://bit.ly/2LomRTp}
\BIBentrySTDinterwordspacing

\bibitem{HarveyPLOSone2013}
E.~S. Harvey, M.~Cappo, G.~A. Kendrick, and D.~L. McLean, ``Coastal fish
  assemblages reflect geological and oceanographic gradients within an
  australian zootone,'' \emph{PLOS ONE}, vol.~8, 2013.

\bibitem{VGG16}
K.~Simonyan and A.~Zisserman, ``Very deep convolutional networks for
  large-scale image recognition,'' \emph{CoRR}, vol. abs/1409.1556, 2014.

\bibitem{ResNet}
K.~He, X.~Zhang, S.~Ren, and J.~Sun, ``Deep residual learning for image
  recognition,'' in \emph{2016 IEEE Conference on Computer Vision and Pattern
  Recognition (CVPR)}, 2016, pp. 770--778.

\bibitem{imagenet_cvpr09}
J.~Deng, W.~Dong, R.~Socher, L.~J. Li, K.~Li, and L.~Fei-Fei, ``Imagenet: A
  large-scale hierarchical image database,'' in \emph{2009 IEEE Conference on
  Computer Vision and Pattern Recognition}, 2009, pp. 248--255.

\bibitem{Razavian2014}
A.~S. Razavian, H.~Azizpour, J.~Sullivan, and S.~Carlsson, ``Cnn features
  off-the-shelf: An astounding baseline for recognition,'' in \emph{2014 IEEE
  Conference on Computer Vision and Pattern Recognition Workshops}, 2014, pp.
  512--519.

\bibitem{MidLevelTrans}
M.~Oquab, L.~Bottou, I.~Laptev, and J.~Sivic, ``Learning and transferring
  mid-level image representations using convolutional neural networks,'' in
  \emph{2014 IEEE Conference on Computer Vision and Pattern Recognition}.\hskip
  1em plus 0.5em minus 0.4em\relax IEEE, 2014, pp. 1717--1724.

\bibitem{TrackingFishSciRep2018}
S.~Marini, E.~Fanelli, V.~Sbragaglia, E.~Azzurro, J.~Del Rio~Fernandez, and
  J.~Aguzzi, ``Tracking fish abundance by underwater image recognition,''
  \emph{Scientific Reports}, vol.~8, p. 13748, 2018.

\bibitem{OBSEA}
\BIBentryALTinterwordspacing
``The western mediterranean expandable seafloor observatory ({OBSEA}).''
  [Online]. Available: \url{http://www.obsea.es}
\BIBentrySTDinterwordspacing

\bibitem{Corgnati2016}
L.~Corgnati, S.~Marini, L.~Mazzei, E.~Ottaviani, S.~Aliani, A.~Conversi, and
  A.~Griffa, ``Looking inside the ocean: Toward an autonomous imaging system
  for monitoring gelatinous zooplankton,'' \emph{Sensors}, vol.~16, 2016.

\bibitem{CLAHE}
K.~Zuiderveld, \emph{"Contrast Limited Adaptive Histogram Equalization"}.\hskip
  1em plus 0.5em minus 0.4em\relax San Diego: "Academic Press Professional",
  1994, pp. 474--485.

\bibitem{Bradley2019}
M.~Bradley, R.~Baker, I.~Nagelkerken, and M.~Sheaves, ``Context is more
  important than habitat type in determining use by juvenile fish,''
  \emph{Landscape Ecology}, 2019, in press.

\bibitem{CIELAB}
\BIBentryALTinterwordspacing
{International Color Consortium}, \emph{Specification ICC.1:2004-10 (Profile
  version 4.2.0.0) Image technology colour management - Architecture, profile
  format, and data structure}, 2004. [Online]. Available:
  \url{http://www.color.org/icc1v42.pdf}
\BIBentrySTDinterwordspacing

\bibitem{WhatIsThePoint2014}
O.~Russakovsky, A.~L. Bearman, V.~Ferrari, and F.~Li, ``What's the point:
  Semantic segmentation with point supervision,'' \emph{CoRR}, vol.
  abs/1506.02106, 2015.

\bibitem{keras}
\BIBentryALTinterwordspacing
F.~Chollet \emph{et~al.}, ``Keras: The python deep learning library,'' 2015.
  [Online]. Available: \url{https://keras.io/}
\BIBentrySTDinterwordspacing

\bibitem{tensorflow2015-whitepaper}
\BIBentryALTinterwordspacing
M.~Abadi \emph{et~al.}, ``{TensorFlow}: Large-scale machine learning on
  heterogeneous systems,'' 2015. [Online]. Available:
  \url{http://tensorflow.org/}
\BIBentrySTDinterwordspacing

\bibitem{Adamax14}
D.~P. Kingma and J.~Ba, ``Adam: {A} method for stochastic optimization,''
  \emph{CoRR}, vol. abs/1412.6980, 2014.

\bibitem{VOC2012}
M.~Everingham, S.~M.~A. Eslami, L.~Van~Gool, C.~K.~I. Williams, J.~Winn, and
  A.~Zisserman, ``The pascal visual object classes challenge: A
  retrospective,'' \emph{International Journal of Computer Vision}, vol. 111,
  pp. 98--136, 2015.

\bibitem{QUT_fish_data}
K.~Anantharajah, Z.~Ge, C.~McCool, S.~Denman, C.~Fookes, P.~Corke,
  D.~Tjondronegoro, and S.~Sridharan, ``Local inter-session variability
  modelling for object classification,'' in \emph{IEEE Winter Conference on
  Applications of Computer Vision}, 2014, pp. 309--316.

\bibitem{QUT2014}
\BIBentryALTinterwordspacing
``{QUT} fish dataset.'' [Online]. Available: \url{https://bit.ly/2APDvGB}
\BIBentrySTDinterwordspacing

\bibitem{WhereAreTheBlobs2018}
I.~H. Laradji, N.~Rostamzadeh, P.~O. Pinheiro, D.~Vazquez, and M.~Schmidt,
  ``Where are the blobs: Counting by localization with point supervision,'' in
  \emph{Computer Vision -- ECCV 2018}, V.~Ferrari, M.~Hebert, C.~Sminchisescu,
  and Y.~Weiss, Eds.\hskip 1em plus 0.5em minus 0.4em\relax Cham: Springer
  International Publishing, 2018, pp. 560--576.

\bibitem{ROC06}
T.~Fawcett, ``An introduction to {ROC} analysis,'' \emph{Pattern Recognition
  Letters}, vol.~27, pp. 861--874, 2006.

\end{thebibliography}

\end{document}